\pdfoutput=1

\documentclass[11pt]{article}

\usepackage{EMNLP2022}

\usepackage{times}
\usepackage{latexsym}
\usepackage{amssymb}
\mathchardef\mhyphen="2D 

\usepackage[T1]{fontenc}

\usepackage[utf8]{inputenc}

\usepackage{microtype}
\usepackage{inconsolata}
\usepackage{float}
\usepackage{multirow}
\usepackage{tablefootnote}
\usepackage{graphicx}
\usepackage{amsmath}

\usepackage{booktabs}

\newcommand{\argmax}{\mathop{\arg\max}}

\newcommand{\Blue}[1]{\textcolor[rgb]{0.00,0.00,1.00}{#1}}

\title{\emph{SubeventWriter}: Iterative Sub-event Sequence Generation with Coherence Controller}

\author{Zhaowei Wang$^1$, Hongming Zhang$^2$, Tianqing Fang$^1$, Yangqiu Song$^1$, \\ \textbf{Ginny Y. Wong$^3$, \& Simon See$^3$}\\
$^1$Department of Computer Science and Engineering, HKUST\\
$^2$Tencent AI Lab, Bellevue, USA\\
$^3$NVIDIA AI Technology Center (NVAITC), NVIDIA, Santa Clara, USA\\
\texttt{\{zwanggy, tfangaa, yqsong\}@cse.ust.hk}\\
\texttt{hongmzhang@global.tencent.com,} \texttt{\{gwong, ssee\}@nvidia.com}}

\begin{document}
\maketitle
\begin{abstract}

In this paper, we propose a new task of sub-event generation for an unseen process to evaluate the understanding of the coherence of sub-event actions and objects. To solve the problem, we design \emph{SubeventWriter}, a sub-event sequence generation framework with a coherence controller. Given an unseen process, the framework can iteratively construct the sub-event sequence by generating one sub-event at each iteration. We also design a very effective coherence controller to decode more coherent sub-events. As our extensive experiments and analysis indicate, \emph{SubeventWriter}\footnote{Code is available at \url{https://github.com/HKUST-KnowComp/SubeventWriter}.} can generate more reliable and meaningful sub-event sequences for unseen processes. 
\end{abstract}

\section{Introduction}
Natural language understanding involves deep understanding of events. In the NLP community, there have been many event understanding tasks.
Most of them focus on parsing events into involved entities, time, and locations as semantic roles~\cite{kingsbury2002treebank,li2013joint,lv2020integrating,DBLP:conf/acl/LinJHW20,du2020event,zhang2021zero,lyu2021zero}, or identifying their binary relations such as temporal or causal relations~\cite{berant-etal-2014-modeling,Event2Mind,Maarten2019Atomic,wang2020joint,wang2021learning}.
However, our natural language can be used to describe relations more than binary ones. For example, processes~\cite{craig1998routledge}, also known as scripts~\cite{schank1977scripts} or activities~\cite{mourelatos1978events}, are complex events constituted by a sequence of sub-events.
Understanding processes can be more challenging than individual or pair of events.

\begin{figure}[t]
    \centering
    \includegraphics[width=\columnwidth]{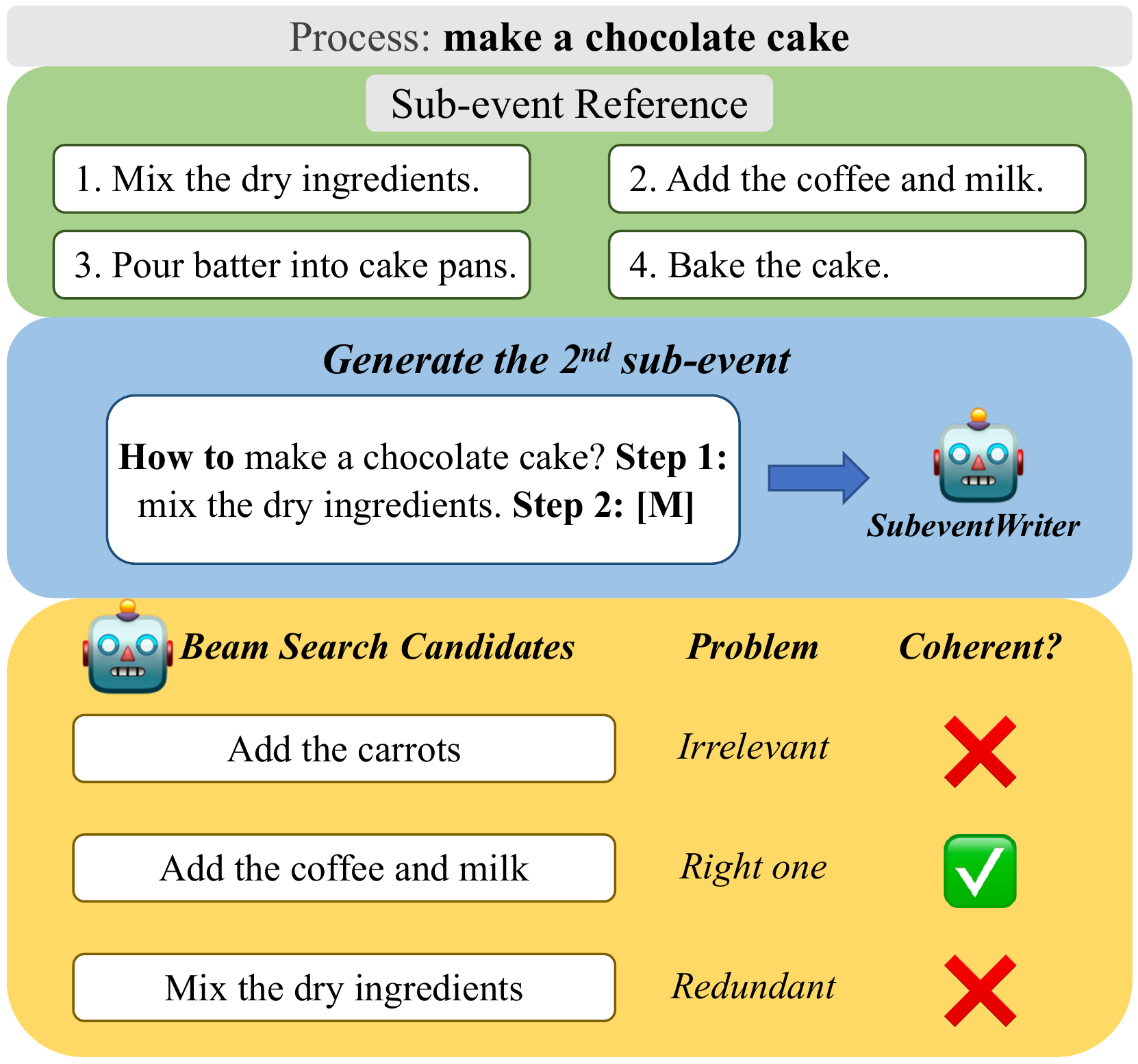}
    \caption{A motivating example of \emph{SubeventWriter}, which generates one sub-event at a time iteratively. We show the process ``make a chocolate cake'' and the second iteration of the generation. By considering coherence, we can re-rank candidates and reach the right sub-event. [M] is a mask token.}
    \label{intro_figure}
\end{figure}


As shown in Figure~\ref{intro_figure}, to complete the process of making a chocolate cake, we need to consider a sequence of actions, ``mix,'' ``add,'' ``pour,'' and ``bake,'' which involves different objects, e.g., dry ingredients, coffee, milk, etc. 
Those actions should follow a logically coherent procedure while the objects should be all related to the target, chocolate cake.
Thus, building such a coherent sequence should take the whole sub-events into consideration.

There have been two categories of related studies to processes, namely process induction and narrative cloze tasks.
\newcite{zhang2020analogous}
proposed a task to learn the hierarchical structure called process induction, where a model needs to generate a sub-event sequence to finish a given process. 
Their framework aggregates existing events so that it can conceptualize and instantiate similar processes.
However, the aggregation procedure does not consider the coherence of actions and their objects.
In addition, to build the dataset, they extracted events using a dependency parser with pre-defined verb-argument templates~\cite{zhang2020aser, zhang2022aser}. Such structured events might harm coherence as only head words are retained after extraction. Consider the first sub-event in Figure~\ref{intro_figure}. After parsing, we lost the indispensable modifier ``dry'' and the sub-event becomes (\emph{mix}, \emph{ingredients})\footnote{The matched pre-defined template is (\emph{verb}, \emph{object}).}, which includes the wet ingredients (e.g., ``milk'') in the second sub-event. Thus, the logical relation between the two adjacent sub-events (i.e., coherence~\cite{van1980semantics}) is defective.
 
On the other hand, narrative cloze tasks~\cite{chambers2008unsupervised,granroth2016happens,chambers2017behind,mostafazadeh2016corpus} evaluate whether a model can predict the missing (usually the last) event in a narrative.
These tasks essentially evaluate the semantic similarity and relatedness between the target event and the context.
However, they did not emphasize how all events in the contexts are unified as a whole process in an ordered and coherent way.

\begin{figure*}[t]
    \centering
    \includegraphics[width=2\columnwidth]{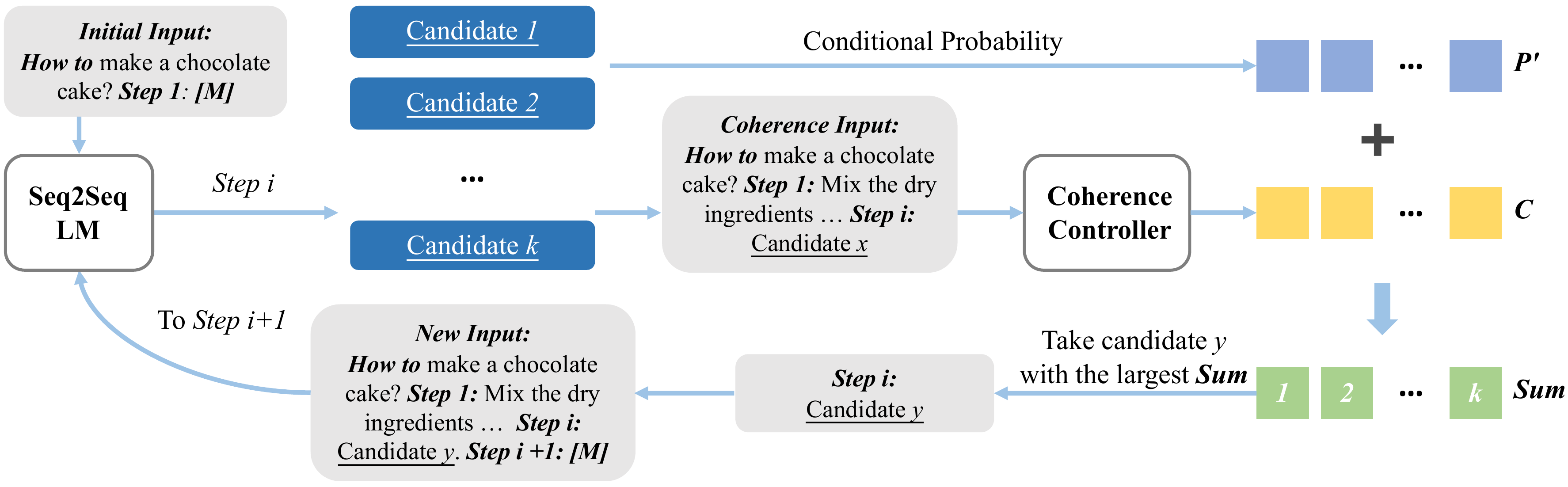}
    \caption{The overview of our \emph{SubeventWriter}. In each iteration, the Seq2Seq language model takes the process and prior generated sub-events as input and generates a few candidates for the next sub-event. Then the coherence controller is used to select the most coherent candidate as the next sub-event.}
    \label{method_figure}
    
\end{figure*}

To evaluate complex process understanding, we propose a new generation-based task to directly generate sub-event sequences in the free-text form, as shown in Figure~\ref{intro_figure}. In the task, better generation of a process means better understanding of the coherence among action verbs as well as their operational objects. In fact, we find that generating free-text events is a non-trivial task, even with existing strong pre-trained models like T5~\cite{raffel2020exploring} and BART~\cite{lewis2020bart}. First, generating an overlong piece of text containing several temporally ordered sub-events at once is challenging to current pre-trained models~\cite{zhou2022claret, lin2021conditional, brown2020language}. Next, sub-events are generated without considering the coherence of actions and their objects, which might give rise to irrelevant or redundant results.

To solve the task, we propose \emph{SubeventWriter} to generate sub-events iteratively in the temporal order. \emph{SubeventWriter} only generates the next sub-event in each generation iteration, given the process and prior generated sub-events. It eases the generation difficulty by decomposing the sub-event sequence. Moreover, sub-events should be coherently organized to complete a process. To consider coherence in each iteration, we can get a few sub-event candidates from the beam search and select the most coherent one, as shown in Figure~\ref{intro_figure}. In \emph{SubeventWriter}, we introduce a coherence controller to score whether a candidate is coherent with the process and prior generated sub-events. As a result, \emph{SubeventWriter} can construct more reliable and meaningful sub-event sequences.
 
To evaluate our framework, we extract a large-scale general-domain process dataset from WikiHow~\footnote{wikihow.com}, containing over 80k examples. We conduct extensive experiments with multiple pre-trained models, and automatic and human evaluations show that \emph{SubeventWriter} can produce more meaningful sub-event sequences compared to existing models by a large margin. Moreover, we conduct few-shot experiments to demonstrate that our framework has a strong ability to handle few-shot cases. Last but not least, we evaluate the generalization ability of \emph{SubeventWriter} on two out-of-domain datasets: SMILE~\cite{regneri2010learning} and DeScript~\cite{wanzare2016crowdsourced}. The results manifest our framework can generalize well.

\section{Textual Sub-event Sequence Generation}
We formally define the sub-event sequence generation task as follows. Given a process $S$, we ask the model to generate sub-event sequences $E$, which are steps to solve the process. This task is essentially a \emph{conditional language modeling} problem. Specifically, given a process $S$ consisting of $n$ tokens: $x_1, x_2, \ldots, x_n$ and a sequence $E$ consists of $m$ sub-events $e_1, e_2, \ldots, e_m$ (each sub-event refers to a sentence containing $t_i$ tokens: $y_{i, 1}, y_{i, 2}, \ldots, y_{i, t_i}$), models aim to learn the conditional probability distribution by maximizing the following conditional probabilities in Eq.~(\ref{language_modeling_equation}):

\begin{equation}
\label{language_modeling_equation}
\begin{gathered}
P_\theta(E|S) = \prod_{i=1}^m P_\theta (e_i|e_{<i},S) \\
P_\theta (e_i|e_{<i},S) = \prod_{j=1}^{t_i} P_\theta (y_{i, j}|y_{i, <j}, e_{<i}, S).
\end{gathered}
\end{equation}

\section{The \emph{SubeventWriter} Framework}
Figure~\ref{method_figure} illustrates the details of the proposed \emph{SubeventWriter} framework. For a given process, the framework decomposes the generation into multiple iterations. The sequence-to-sequence (seq2seq) language model generates a few candidates for the next sub-event in each iteration. We then leverage a coherence controller to re-rank the generated candidates by considering whether they are coherent with the process and prior generated sub-events. The coherence controller is a discriminative model that can assign a coherence score to a sub-event sequence. It is fine-tuned independently on our synthetic data generated according to our manually designed coherence rules. Finally, the framework appends the generated sub-event to the end of the input to serve as new context and start the next iteration. The detailed description of \emph{SubeventWriter} components is as follows:

\subsection{Iterative Event-level Decoding}
The iterative event-level decoding scheme is built on top of seq2seq language models, including T5~\cite{raffel2020exploring} and BART~\cite{lewis2020bart}. We describe training and inference details as follows.

\textbf{Training:} 
The seq2seq language models are fine-tuned to decode one sub-event each time in chronological order. 
For each process with its sub-event sequences in the training data, we create an augmented set of training examples with each sub-event in the sequence as the output in turns. For example, if the valid sequence of a process $S$ consists of temporally ordered sub-events $e_1$, $e_2$, and $e_3$, we then create four training examples: $S \rightarrow e_1$, $S \cup \{e_1\} \rightarrow e_2$, $S \cup \{e_1, e_2\} \rightarrow e_3$, and $S \cup \{e_1, e_2, e_3\} \rightarrow \textit{none}$, where ``$\textit{none}$'' is a special token to end sequences. The order of adding sub-events $e_i$ follows the temporal order, which ensures that the model only needs to predict what will happen next without a longer-term forecast.

To minimize the gap between pre-training and fine-tuning, we design a textual prompt template to construct input in human language. If we want to generate the $i+1$ \emph{th} sub-event given a process $S$ and sub-events $e_1, e_2, ..., e_i$, the template takes the form of ``How to $S$? Step 1: $e_1$ ... Step i: $e_i$. Step i+1: [M]'' as the example shown in Figure~\ref{method_figure}. [M] is the mask token of the model. More examples of input/output are shown in Appendix~\ref{appendix_prompt_io_example}.

\textbf{Inference:}
During the inference, we apply the seq2seq language models iteratively to generating the sub-event sequence of a process. The aforementioned prompt template is also used. For instance, the model first generates sub-event $e_1$ for a process $S$. It then takes $S$ and $e_1$ as input and generates the second sub-event $e_2$. The model repeats this process until the special token ``$\textit{none}$'' is generated, which means no more sub-events are required. Then, generated sub-events are concatenated into a sequence as the final output.

\subsection{Coherence Controller}
As a sub-event sequence should be coherent to complete a process, we propose a coherence controller to control the iterative event-level decoding. At each iteration, the coherence controller considers whether each sub-event candidate is coherent with the given process and sub-events generated in previous iterations. Considering that sub-events (one or more sentences) are diverse and complicated, here we employ a coherence model~\cite{jwalapuram2021rethinking} based on BERT~\cite{devlin2018bert} as the coherence controller to score sub-event candidates.

We train the coherence controller as a binary classification task to discriminate coherent sub-event sequences from incoherent ones. Following previous works~\cite{mesgar2018neural, moon2019unified}, we regard a human-written sub-event sequence as coherent, and we synthetically build two types of incoherent sub-event sequences by corrupting the local or global coherence of the human-written one. For \textbf{\emph{local coherence}}, we randomly copy a sub-event in the current process and place this \textbf{\emph{duplicate sub-event}} at a random location. In this way, the relation between two sub-events adjacent to the duplicate sub-event is corrupted, entitled local coherence in linguistics~\cite{van1980semantics}. For \textbf{\emph{global coherence}}, we randomly choose a sub-event from other processes with a different theme and insert this \textbf{\emph{irrelevant sub-event}} at a random location. In this way, the theme among all sub-events is corrupted, called global coherence~\cite{van1980semantics}. We show positive and two types of negative examples in Appendix~\ref{appendix_training_coherence_controller_example}.

We use the cross-entropy loss shown in Eq.~\ref{coherence_loss_equation} to optimize the coherence controller, where $y$ and $\hat{y}$ are label and coherence scores, respectively. Since $y$ equals $1$ for positive examples, our model will give higher scores for more coherent input. For each positive example, we sample $N$ negative examples by corrupting local coherence and the same number by corrupting global coherence ($2N$ in total). Thus, we balance the loss function by dividing negative loss by $2N$:
\begin{equation}
\label{coherence_loss_equation}
\begin{gathered}
    \mathcal{L}_{coh} = - (y \log \hat{y} + \frac{1}{2N}(1-y) \log (1-\hat{y})).
\end{gathered}
\end{equation}

At the inference stage (Figure~\ref{method_figure}), we concatenate the process and the current generated sequence into the input to the coherence controller. For example, in the $i$ \emph{th} iteration, the Seq2Seq language model with beam search
returns top-$k$ possible sub-event candidates: $\hat{e}_{i1}, \hat{e}_{i2}, \ldots, \hat{e}_{ik}$. We construct the input $S; \hat{e}_1, \hat{e}_2, \ldots, \hat{e}_{i-1}, \hat{e}_{ij}$ for every candidate $\hat{e}_{ij}$, given process $S$ and prior sub-events $\hat{e}_1, \hat{e}_2, \ldots, \hat{e}_{i-1}$. With such input, the coherence controller computes coherence scores $C(\hat{e}_{ij})$. As the sequence-to-sequence model can return the logarithm of conditional probability $P' (\hat{e}_{ij}) = \log P_\theta (\hat{e}_{ij}|\hat{e}_{<i},S)$ (Eq.~\ref{language_modeling_equation}) for each candidate, we re-rank candidates and return the best one according to the sum of the two scores:
\begin{equation}
\label{sum_score_equation}
\begin{gathered}
    \hat{e}_i = \argmax_{\hat{e}_{ij}}\ \{P' (\hat{e}_{ij}) + \lambda C(\hat{e}_{ij})\},
\end{gathered}
\end{equation}
where $\lambda$ is a hyper-parameter to weight coherence scores. Appendix~\ref{appendix_inference_coherence_controller_example} gives a concrete example of the inference stage of the coherence controller.

\section{Experiments}
We conduct extensive experiments and compare \emph{SubeventWriter} with a wide selection of baselines.

\subsection{Dataset}
We collect processes and corresponding event sequences from the WikiHow website\footnote{wikihow.com}~\cite{koupaee2018wikihow}, where each process is associated with a sequence of temporally ordered human-annotated sub-events. We randomly split them into the training, validation, and testing sets. As a result, we got 73,847 examples for the training set and 5,000 examples for both validation and testing sets, whose average sub-event sequence length is 4.25. 

\subsection{Evaluation Metric}
For each pair of a predicted sequence and a ground truth sequence, we compute BLEU-1~\cite{papineni2002bleu}, BLEU-2, ROUGE-L~\cite{lin2004rouge}, and BERTScore~\cite{zhang2019bertscore} between them and take the average of each metric over all data. For inference cases with multiple references, we take the best performance among all references. 

\begin{table*}[t]
    \small
	\centering
	
	\begin{tabular}{l||cccc|cc}
	    \toprule
		\textbf{Models}&\textbf{B-1}&\textbf{B-2} &\textbf{R-L}&\textbf{BERT} & \textbf{$\Delta_{B\mhyphen1}$} & \textbf{$\Delta_{B\mhyphen2}$}\\
		\midrule
		Zero-shot Large LM (GPT-J 6b) &13.88&0.33&16.47&45.43 & - & -\\
		Zero-shot Large LM (T5-11b) &20.14&0.76&14.11&54.55 & - & - \\
		\midrule
		Top-1 Similar Sequence (Glove) &16.31&0.99 &11.63&57.24& - & - \\
		Top-1 Similar Sequence (\textsc{SBert})  &18.39&2.21 &13.46&59.94& - & - \\
		\midrule
        All-at-once Seq2Seq (BART-base) &21.01&4.52 &18.83&58.79& - & - \\
        All-at-once Seq2Seq (BART-large) &21.84&4.73 &18.94&59.45& - & -  \\
		\midrule
        All-at-once Seq2Seq (T5-base) &20.33&5.63& 20.22&52.15& - & -  \\
        All-at-once Seq2Seq (T5-large) &24.27&7.11&21.76&57.58& - & -  \\
        All-at-once Seq2Seq (T5-3b) &27.99&8.72&23.36&62.03& - & -  \\
        \midrule
        \emph{SubeventWriter} (BART-base) &29.62&8.35 &21.59&60.42& $\uparrow$ 8.61 & $\uparrow$ 3.83\\
        \emph{SubeventWriter} (BART-large) &31.31&9.41 &22.52&61.83& $\uparrow$ 9.47 & $\uparrow$ \textbf{4.68}\\
        \midrule
        \emph{SubeventWriter} (T5-base) &30.74&8.89&22.44&61.81& $\uparrow$ \textbf{10.41} & $\uparrow$ 3.26\\
        \emph{SubeventWriter} (T5-large) &33.01&10.39&23.07&64.19&  $\uparrow$ 8.74 & $\uparrow$ 3.28\\
        \emph{SubeventWriter} (T5-3b) &\textbf{34.75}&\textbf{11.30}&\textbf{24.17}&\textbf{65.67}& $\uparrow$ 6.76 & $\uparrow$ 2.58\\
		\bottomrule
	\end{tabular}
	\caption{Performance of all frameworks on the testing set of the WikiHow dataset. \emph{SubeventWriter} is our model. We abbreviate BLEU-1, BLEU-2, ROUGE-L, and BERTScore to B-1, B-2, R-L, and BERT, respectively. Compared to All-at-once Seq2Seq, improvements of our frameworks are shown under \textbf{$\Delta_{B\mhyphen1}$} and \textbf{$\Delta_{B\mhyphen2}$} for each size of T5 and BART. We also include the performance of all models on the validation set in Appendix~\ref{appendix_wikihow_valiation}.}
	\label{wikihow_test}
\end{table*}

\subsection{Baseline Methods}
\label{section_baseline_methods}
We compare our framework to three methods:\\
\textbf{All-at-once Seq2Seq:} An intuitive solution to the textual sub-event sequence generation task would be modeling it as an \emph{end-to-end sequence-to-sequence (Seq2Seq) problem}, where Seq2Seq language models are fine-tuned to predict all sub-events at once, given a process as input. Here we test multiple Seq2Seq language models: T5-base/large/3b and BART-base/large. We refer to this baseline as ``All-at-once'' for short in following sections.

\noindent\textbf{Top-one Similar Sequence:} Following previous work~\cite{zhang2020analogous}, another naive yet potentially strong baseline is Top-one Similar Sequence. For each unseen process in the validation or testing set, the baseline finds the most similar process in the training data. The sub-event sequence of the most similar process is then regarded as the prediction. If more than one sub-event sequence exists for the most similar process, we randomly pick one from them. Here, we consider two methods to measure similarities: cosine similarity of Glove~\cite{pennington2014glove} and Sentence-BERT (\textsc{SBert})~\cite{reimers2019sentence} embeddings.

\noindent\textbf{Zero-shot Large LM:} 
Large language models (LMs) have shown stronger performance on extensive NLP tasks~\cite{raffel2020exploring}. The third baseline we introduce is prompting large language models in the zero-shot setting. We consider GPT-J~\cite{gpt-j} and T5-11b, which contain \textasciitilde 6 billion and \textasciitilde 11 billion parameters, respectively. We choose the prompt template ``How to \emph{S}? Generate the events to solve it.'' for every process \emph{S}. 

\subsection{Implementation Details}
We fine-tune \emph{SubeventWriter} and All-at-once Seq2Seq based on T5-base/large/3b and BART-base/large for four epochs. The best checkpoint is selected according to the sum of all metrics on the validation set. The grid search explored learning rates of 1e-5, 5e-5, 1e-4, 5e-4, batch size of 32 and 64, and weight $\lambda$ of coherence scores (Eq.~\ref{sum_score_equation}) of 0.5, 1, 2, 5. We test multiple LMs for the coherence controller and choose BERT-base due to its efficiency. We show more details in Appendix~\ref{appendix_implementation}.

\section{Main Evaluation}
\label{main_evaluation_section}
We show the results on the testing set of the WikiHow dataset in Table~\ref{wikihow_test}. In general, \emph{SubeventWriter} can generate relevant sub-event sequences, outperforming all baseline frameworks by a great margin. For example, 11.30\% of bi-grams generated by \emph{SubeventWriter} (T5-3b) are covered by the references, increasing by 2.58\% absolutely and 29.6\% relatively compared to All-at-once Seq2Seq (T5-3b). Even though GPT-J and T5-11b are much larger, the smallest fine-tuned \emph{SubeventWriter} (BART-base) can still surpass them.

Besides, our framework improves more significantly on smaller-sized language models and is parameter efficient. With T5 going down from ``3b'' to ``base'', we observe improvements increase (e.g., from 6.76\% to 10.41\% for BLEU-1). Also, \emph{SubeventWriter} (T5-base) achieves comparable performance compared to All-at-once Seq2Seq (T5-3b) with only about 12\% parameters\footnote{We include the parameters of both BERT-base in the coherence controller and T5-base.} because generating one event each time is not hard to T5-base with the help of the coherence controller.

Another interesting observation is that when comparing between \emph{SubeventWriter} based on T5 and BART, T5 always performs slightly better than BART in both ``base'' and ``large'' sizes. Such advances are consistent with intuition since T5 is about 1.5x - 2x larger than BART.

In the rest of this section, we conduct more analysis to demonstrate the reason behind the success of \emph{SubeventWriter}.

\subsection{Ablation Study}
To measure the contribution of each module to the final results, we conduct an ablation study on \emph{SubeventWriter} in Table~\ref{ablation_study_table}.

The first ablation experiment drops the coherence controller in \emph{SubeventWriter} ($\diamond$ w/o CoCo), which verifies the effectiveness of controlling coherence. Then, the second ablation experiment further drops iterative event-level decoding ($\diamond$ w/o CoCo \& ITER.) and substantiates the iterative event-level decoding can boost the performance. The coherence controller depends on iterative event-level decoding as it controls sub-events one by one in chronological order. Thus, we cannot only drop iterative event-level decoding while the coherence controller is kept (no $\diamond$ w/o ITER.).

From the results in Table~\ref{ablation_study_table}, we observe that both the coherence controller and the iterative event-level decoding play essential roles in generating high-quality sub-event sequences. Dropping each of them will cause drastic decreases in all metrics. Taking \emph{SubeventWriter} (BART-large) as an example, the BLEU-1 decreases by 4.53\% without the coherence controller. When the iterative event-level decoding is further removed, the BLEU-1 declines to 21.84\%, which is only 70\% of the original BLEU-1 score.
 
\begin{table}[t]
    \small
	\centering
	\begin{tabular}{l|cccc}
	    \toprule
		\textbf{Models}&\textbf{B-1}&\textbf{B-2}&\textbf{R-L}&\textbf{BERT}\\
		\midrule
		Ours (BART-large) &\textbf{31.31}&\textbf{9.41}&\textbf{22.52}&\textbf{61.83} \\
		\midrule
        $\diamond$ w/o CoCo &26.78&7.79&22.04&59.53\\
		$\diamond$ w/o CoCo \& ITER. &21.84&4.73 &18.94&59.45\\
		\midrule \midrule
		Ours (T5-large) &\textbf{33.01}&\textbf{10.39}&\textbf{23.07}&\textbf{64.19}\\
		\midrule
        $\diamond$ w/o CoCo &30.41&9.14&22.75&62.19\\
		$\diamond$ w/o CoCo \& ITER. &24.27&7.11&21.76&57.58\\
		\bottomrule
	\end{tabular}
	\caption{Ablation study on \emph{SubeventWriter}. ``w/o CoCo'' refers to ablation of coherence controller. Further ablation of iterative event-level decoding is shown in ``w/o CoCo \& ITER.'' See Appendix~\ref{appendix_ablation_study} for results on other sizes of BART and T5.}
	\label{ablation_study_table}
\end{table}
 
\begin{figure}[t]
    \centering
    \includegraphics[width=\columnwidth]{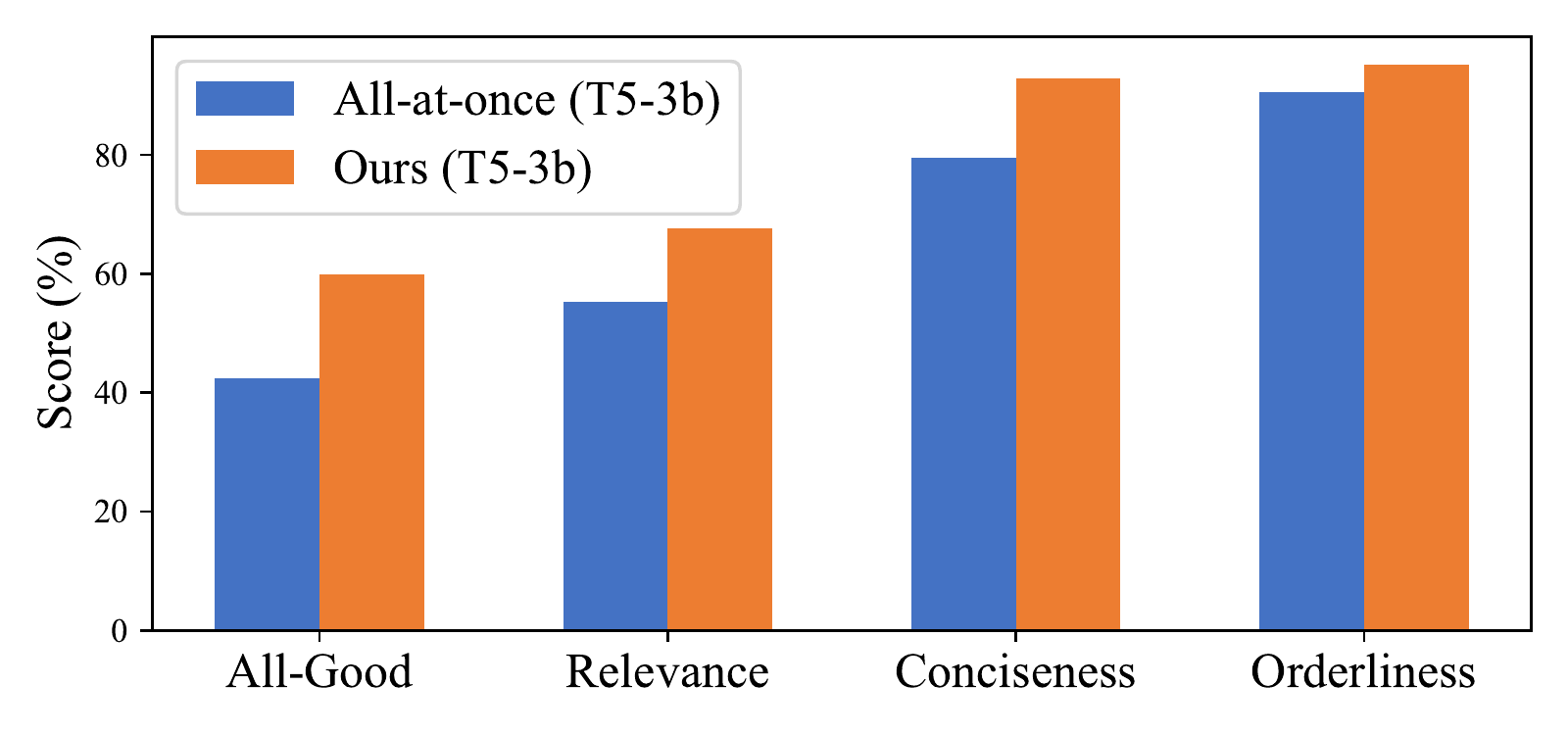}
    \caption{Human evaluation scores of \emph{SubeventWriter} (``Ours'') and All-at-once Seq2Seq (``All-at-once'') based on T5-3b. ``All-Good'' means sub-events that satisfy all three aspects.}
    \label{human_eval_figure}
\end{figure} 
 
\subsection{Human Evaluation}
We perform human evaluation to complement automatic evaluation. As sub-event sequences are complicated and diverse, we decompose them and score every sub-event in the following 3 aspects: 

\noindent\textbf{Relevance:} whether a sub-event is relevant to solving the given process, measuring how well sub-events focus on the same theme (global coherence).

\noindent\textbf{Conciseness:} whether a sub-event is not redundant to others in the same sequence. We introduce this aspect since generating duplicates is a common failure of language models~\cite{brown2020language} and destroys local coherence.

\noindent\textbf{Orderliness:} whether a sub-event is placed in proper order, considering its prior sub-events. As the order of sub-events irrelevant to the given process is not defined clearly, we only consider the order of sub-events that satisfy the first aspect (Relevance).

We choose to evaluate the generation of \emph{SubeventWriter} (T5-3b) and All-at-once (T5-3b) as they have the best quantitative performance. We randomly select 50 processes from the testing set, containing about 200 sub-events. Three experts are asked to evaluate every sub-event, yielding 1,800 total ratings for each model (200 sub-events $\times$ 3 aspects $\times$ 3 experts). We take the majority vote among three votes as the final result for each sub-events. The IAA score is 83.78\%
calculated using pairwise agreement proportion, and the Fleiss’s $\kappa$~\cite{fleiss1971measuring} is 0.57.

We show the average scores in Figure~\ref{human_eval_figure}. We can observe that both models achieve acceptable scores in orderliness. \emph{SubeventWriter} generates more relevant and less redundant sub-events in comparison to All-at-once Seq2Seq as the global and local coherence in the coherence controller reflects the relevance and conciseness, respectively.
Our model also produces more ``All-Good'' sub-events, which satisfy all three aspects. 

\begin{figure}[t]
    \centering
    \includegraphics[width=\columnwidth]{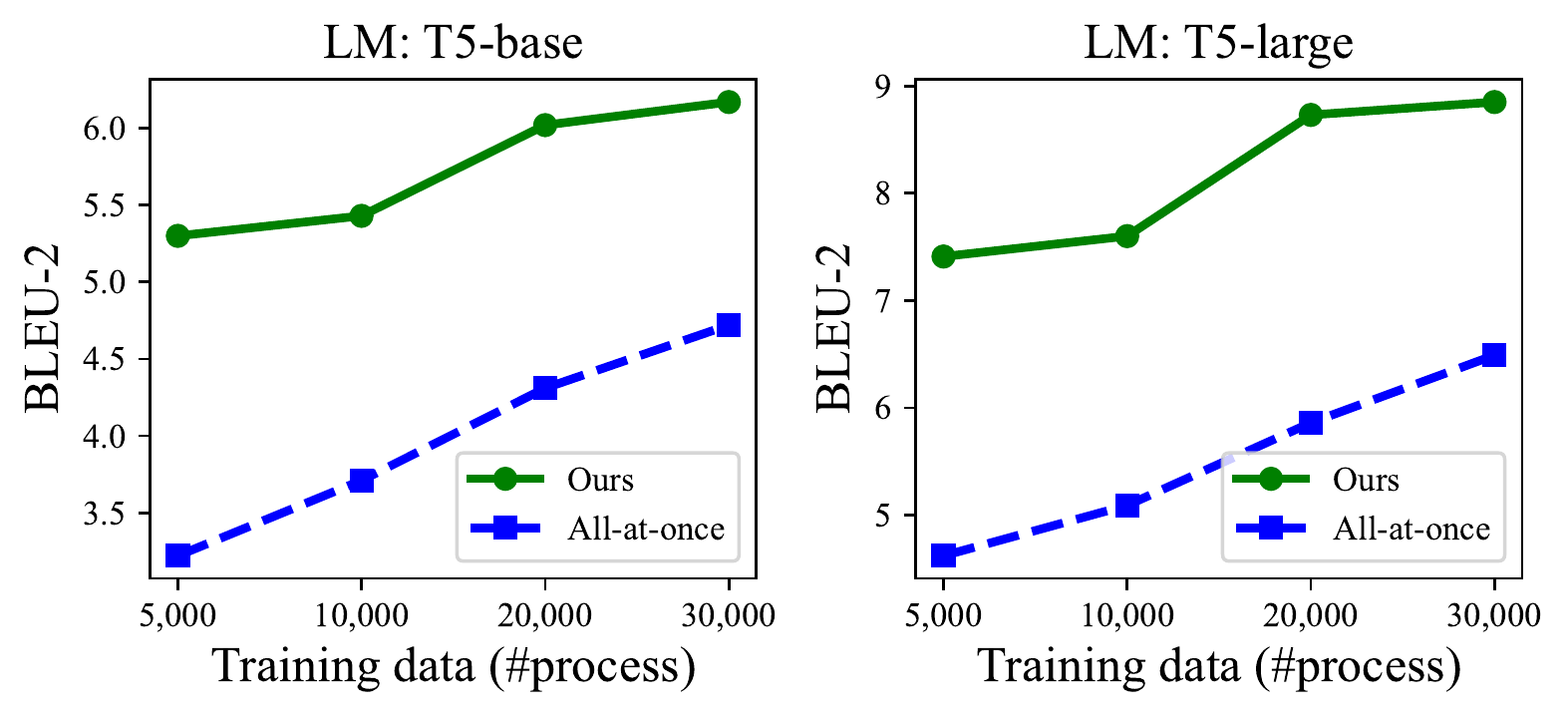}
    \caption{Few-shot learning performance of \emph{SubeventWriter} (``Ours'') and All-at-once Seq2Seq (``All-at-once'') based on T5 is shown. We also include the results of other metrics in Appendix~\ref{appendix_few_shot}.}
    \label{few_shot_figure}
\end{figure}

\begin{table*}[t]
    \small
	\centering
	\begin{tabular}{l|cccc|cccc}
	    \toprule
	    \multirow{2}{*}{\textbf{Models}} &\multicolumn{4}{c|}{SMILE} &\multicolumn{4}{c}{DeScript} \\ \cmidrule(lr){2-5} \cmidrule(lr){6-9}
		&\textbf{B-1}&\textbf{B-2}&\textbf{R-L}&\textbf{\textsc{Bert}} &\textbf{B-1}&\textbf{B-2}&\textbf{R-L}&\textbf{\textsc{Bert}}\\
		\midrule
		Top-1 Similar Sequence (Glove) &24.42&0.91&12.54&52.66
		&39.26&5.50&16.30&56.80\\
		Top-1 Similar Sequence (SBERT) &25.98&1.85&11.62&53.28
		&45.89&8.57&17.02&58.73\\
	    \midrule
		All-at-once  (BART-base) &42.22&10.14&20.51&54.49 &66.90&28.61&28.35&60.99\\
        All-at-once (BART-large) &46.25&11.45&21.80&55.79 &71.04&33.27&29.29&62.15\\
        \midrule
        All-at-once (T5-base) &29.07&5.56&22.57&45.38 &59.07&20.43&32.53&54.69\\
        All-at-once (T5-large) &37.31&11.07&25.54&55.69 &62.97&27.57&33.69&59.91\\
        All-at-once (T5-3b) &40.11&10.23&26.41&60.29 &74.60&44.15&36.14&68.50\\
        \midrule
        Ours (BART-base) &44.67&10.83&24.07&54.19 &74.02&36.93&31.03&61.69\\
        Ours (BART-large) &48.78&16.22&27.17&56.53 &80.95&43.93&33.70&64.39\\
        \midrule
        Ours (T5-base) &43.71&10.44&24.09&56.90
        &76.41&39.74&32.73&63.84\\
        Ours (T5-large)  &45.13&12.68&26.51&60.29 &82.54&47.34&35.29&68.32\\
        Ours (T5-3b) &\textbf{53.41}&\textbf{15.21}&\textbf{28.69}&\textbf{60.68} &\textbf{86.27}&\textbf{51.24}&\textbf{37.10}&\textbf{70.17}\\
		\bottomrule
	\end{tabular}
	\caption{Performance of zero-shot transfer learning on SMILE and DeScript. \emph{SubeventWriter} (``Ours'') outperforms the All-at-once Seq2Seq baseline (``All-at-once'') by a large margin.}
	\label{smile_descript_zero_shot_table}
\end{table*}

\begin{table}[t]
    \small
    \centering
	\begin{tabular}{l|cccc}
	    \toprule
		\textbf{Dataset}&\textbf{\#Process}&\textbf{\#Seq}&\textbf{Avg-Ref}&\textbf{Avg-Len}\\
		\midrule
		SMILE & 22 & 386 & 17.55 & 9.06\\
		DeScript & 42 & 3845 & 91.55 & 8.32\\
		\bottomrule
	\end{tabular}
	\caption{Statistics of SMILE and DeScript. \textbf{\#Process}, \textbf{\#Seq}, \textbf{Avg-Ref}, and \textbf{Avg-Len} are the number of processes, sub-event sequences, average sequences per process, and average sub-events per sequence, respectively.}
	\label{OOD_stat}
\end{table}

\subsection{Few-shot Learning Ability}
We conduct main evaluation on the WikiHow dataset, which contains a large training set. To better understand the generalization ability of \emph{SubeventWriter}, we conduct few-shot experiments to confirm its ability to generalize with fewer data.

Referring to the size of the validation set (5,000 examples), we conduct experiments with training data of 5,000, 10,000, 20,000, and 30,000 shots (1x, 2x, 4x, 6x as large as the validation set)\footnote{The full training set is 15x as large as the validation set}. As shown in Figure~\ref{few_shot_figure}, \emph{SubeventWriter} achieves better performance compared to All-at-once Seq2Seq, demonstrating that \emph{SubeventWriter} owns the ability to generalize with fewer data.

\subsection{Zero-shot Transfer Learning}
To further verify the generalization ability of \emph{SubeventWriter}, we test it on two small-scale and domain-specific datasets: SMILE~\cite{regneri2010learning} and DeScript~\cite{wanzare2016crowdsourced}. Both contain hundreds of human-curated sub-event sequences pertaining to human activities. Statistics for each dataset are in Table~\ref{OOD_stat}. 

We directly use \emph{SubeventWriter} fine-tuned on the WikiHow dataset to test its zero-shot transferring ability because it performs well on the WikiHow dataset.
Since we do not tune hyper-parameters on these datasets, we treat each entire dataset as a testing set, and there is no validation set. 

We report the zero-shot transferring results on \emph{SubeventWriter} on SMILE and DeScript in Table~\ref{smile_descript_zero_shot_table}. Among all baseline methods introduced in Section~\ref{section_baseline_methods}, we choose Top-one Similar Sequence and All-at-once Seq2Seq as baselines. The method Zero-shot Large LM does not fit WikiHow data, so it is not suitable to test the zero-shot transferring ability. From Table~\ref{smile_descript_zero_shot_table}, we can find the performance on SMILE and DeScript is higher than the WikiHow dataset for all models since more references are provided. We find \emph{SubeventWriter} surpasses Top-one Similar Sequence and All-at-once Seq2Seq on both datasets and all model sizes. Such improvements indicate that our framework is able to learn non-trivial knowledge about sub-event sequences and has a strong generalization ability.


\subsection{Cutting Down the Model Parameters}
Most of our experiments use T5-base (\textasciitilde 220M parameters) and T5-large (\textasciitilde 770M parameters), or the counterpart of BART, but in practice, we might prefer to use smaller models due to computational limitations. Here, we investigate the impact of model size by using T5-small (\textasciitilde 60M parameters) models. Table~\ref{small_size_t5} presents the results for fine-tuning All-at-once Seq2Seq (T5-small) and \emph{SubeventWriter} (T5-small). Since the coherence controller is based on BERT-base (\textasciitilde 110M parameters), we remove it from \emph{SubeventWriter} to keep the number of parameters consistent for a fair comparison. 

There are two meaningful observations. First, \emph{SubeventWriter} (T5-small) can still provide superior sub-event sequences compared to All-at-once Seq2Seq (T5-small) by a large margin (e.g., 6.35\% in BLEU-1). Second, both \emph{SubeventWriter} and All-at-once Seq2Seq perform worse when we replace T5-base with T5-small since the model size reduces to 27\% of the original one. 

\begin{table}[t]
    \small
	\centering
	\begin{tabular}{l|cccc}
	    \toprule
		\textbf{Models}&\textbf{B-1}&\textbf{B-2}&\textbf{$\delta_{B\mhyphen1}$}&\textbf{$\delta_{B\mhyphen2}$}\\
		\midrule
        All-at-once & 17.18 & 4.25 & $\downarrow$ 3.15 & $\downarrow$ 1.38\\
        \emph{SubeventWriter} & \textbf{23.53} & \textbf{5.67} & $\downarrow$ 5.03 & $\downarrow$ 2.60 \\
		\bottomrule
	\end{tabular}
	\caption{Performance of using T5-small. \textbf{$\delta_{B\mhyphen1}$} and \textbf{$\delta_{B\mhyphen2}$} indicate performance drops when replacing T5-base with T5-small. See Appendix~\ref{appendix_t5_small} for full results.}
	\label{small_size_t5}
\end{table}

\subsection{Comparison of Sub-event Sequence Length}
While prior experiments and analysis mainly focus on the generated content, we also compare lengths of generated sub-event sequences with the ground truth to better assess \emph{SubeventWriter}. We consider the lengths as a regression problem and decide to use two metrics: Mean Absolute Error (MAE) and Root Mean Squared Error (RMSE).

From Table~\ref{sequence_length_table}, we can observe that \emph{SubeventWriter} achieves less mean absolute and root mean squared error than All-at-once Seq2Seq, which indicates that our framework can generate sequences with more precise numbers of sub-events.

\begin{table}[t]
    \small
	\centering
	\setlength{\tabcolsep}{4pt}
	\begin{tabular}{l|cccc}
	    \toprule
	    \multirow{2}{*}{\textbf{Models}} &\multicolumn{2}{c}{\textbf{Valid}} &\multicolumn{2}{c}{\textbf{Test}} \\  \cmidrule(lr){2-3} \cmidrule(lr){4-5}
		&MAE&RMSE&MAE&RMSE\\
		\midrule
		 All-at-once (BART-large) &2.25&2.60&2.27&2.64\\
        All-at-once (T5-large) &1.37&1.88&1.39&1.89 \\
        \midrule
        Ours (BART-large) &1.40&1.86&1.45&1.90\\
        Ours (T5-large) &\textbf{1.34}&\textbf{1.80}&\textbf{1.38}&\textbf{1.84}\\
		\bottomrule
	\end{tabular}
	\caption{Regression errors of sub-event sequence length of \emph{SubeventWriter} and All-at-once Seq2Seq, which verify our framework can predict precise lengths.}
	\label{sequence_length_table}
\end{table}

\subsection{Case Study}
\begin{figure}[t]
    \centering
    \includegraphics[width=\columnwidth]{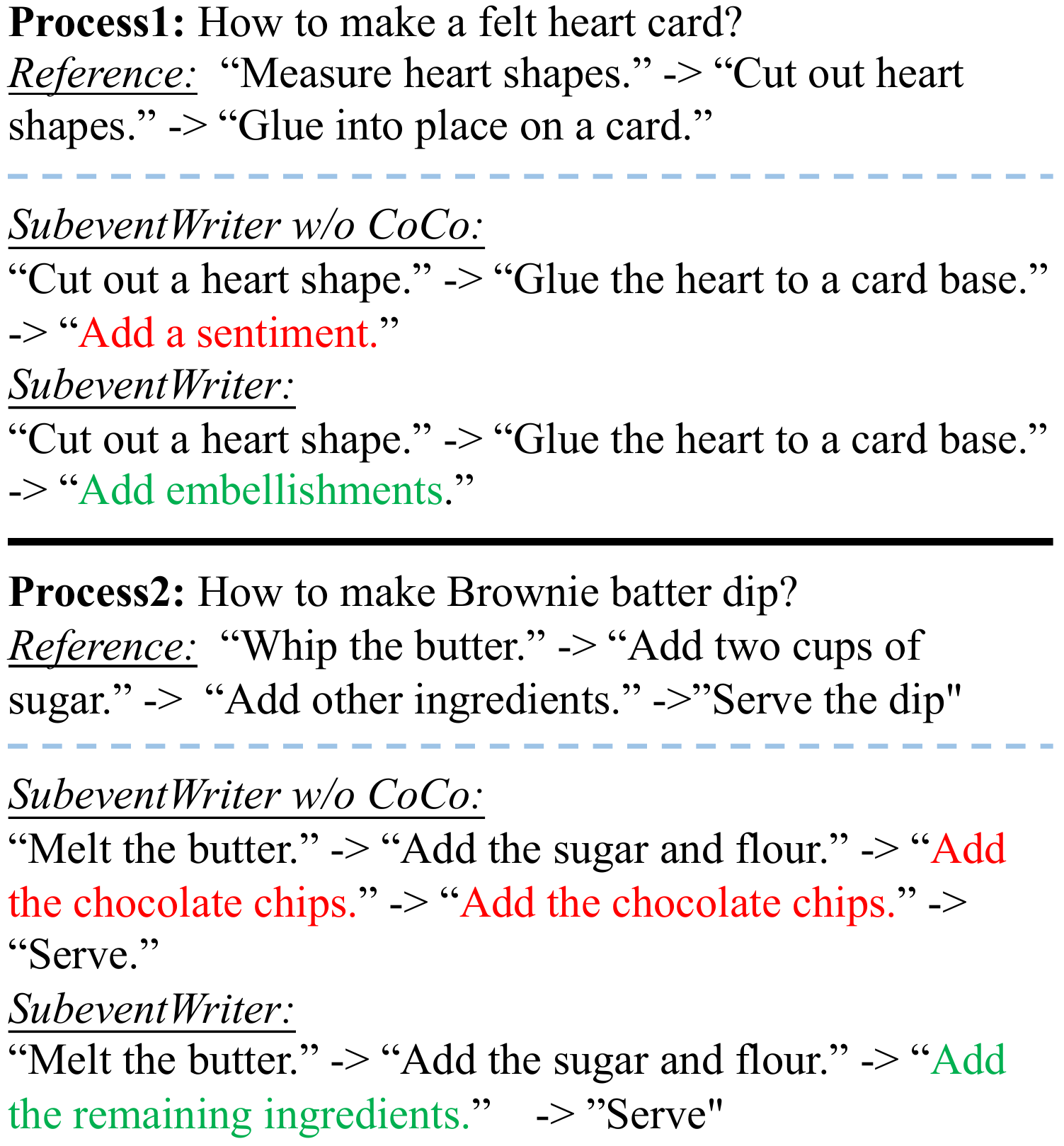}
    \caption{Case Study. We show the generation without coherence controller (w/o CoCo) to illustrate how \emph{SubeventWriter} works. We mark digressive (in Process1) and redundant (in Process2) sub-events with red. Their corrections are marked with green.}
    \label{case_study_figure}
\end{figure}

We show two sub-event sequences produced by \emph{SubeventWriter} in Figure~\ref{case_study_figure}. We also generate sub-events without the coherence controller to analyze how it works. From the first example, we can observe that \emph{SubeventWriter} without the coherence controller produces a digressive sub-event ``Add a sentiment.'' The coherence controller can correct the generation and keep a consistent theme across the process and all sub-events (global coherence). From the second example, two redundant sub-events, ``Add the chocolate chips,'' are generated with flawed discourse relation between them (local coherence). We can see the coherence controller rectifies the redundancy by considering coherence in decoding. 

\section{Related Work}
Understanding events has been a challenging task in NLP for a long time~\cite{chen2021event}, to which the community has dedicated many works. \newcite{chambers2008unsupervised} first introduced the narrative cloze task, where models are asked to predict the next event from given ones. After them, a few works are devoted to better modeling the event representations~\cite{pichotta2014statistical, pichotta2016learning, granroth2016happens, li2018constructing, ding2019event, bai2021integrating}. \newcite{mostafazadeh2016corpus} studied the story cloze test, where a system needs to choose the correct ending for a short story. Nonetheless, those tasks emphasize the semantic similarity and relatedness among events, ignoring how events are organized coherently.

A similar work to ours is process induction~\cite{zhang2020analogous}, where they proposed a statistical framework to generate a sub-event sequence of a given process. The framework aggregates existing events with conceptualization and instantiation. The difference between our work and theirs is that we consider the coherence of both actions and their objects in generation. Tasks about processes in different forms are also studied, including sub-event sequence typing~\cite{chen2020you, pepe2022steps}, sub-event selection~\cite{zhang2020reasoning}, chronological ordering~\cite{jin2022probing}, script construction with specified length~\cite{lyu2021goal}, and multi-relation prediction~\cite{lee2019multi}. Compared to their settings, our work directly tackles the most challenging one, where models are asked to generate whole sub-event sequences.

\section{Conclusion}
In this paper, we try to construct coherent sub-event sequences by considering coherence in event-level decoding. Our \emph{SubeventWriter} generates sub-events iteratively. A coherence controller is introduced to re-rank candidates in each iteration. The extensive experiments demonstrate the effectiveness of \emph{SubeventWriter}.

\section{Acknowledgement}
The authors of this paper were supported by the NSFC Fund (U20B2053) from the NSFC of China, the RIF (R6020-19 and R6021-20) and the GRF (16211520) from RGC of Hong Kong, the MHKJFS (MHP/001/19) from ITC of Hong Kong and the National Key R\&D Program of China (2019YFE0198200) with special thanks to HKMAAC and CUSBLT, and  the Jiangsu Province Science and Technology Collaboration Fund (BZ2021065). We also thank the support from NVIDIA AI Technology Center (NVAITC) and the UGC Research Matching Grants (RMGS20EG01-D, RMGS20CR11, RMGS20CR12, RMGS20EG19, RMGS20EG21).

\section*{Limitations}
\begin{figure}[t]
    \centering
    \includegraphics[width=0.8\columnwidth]{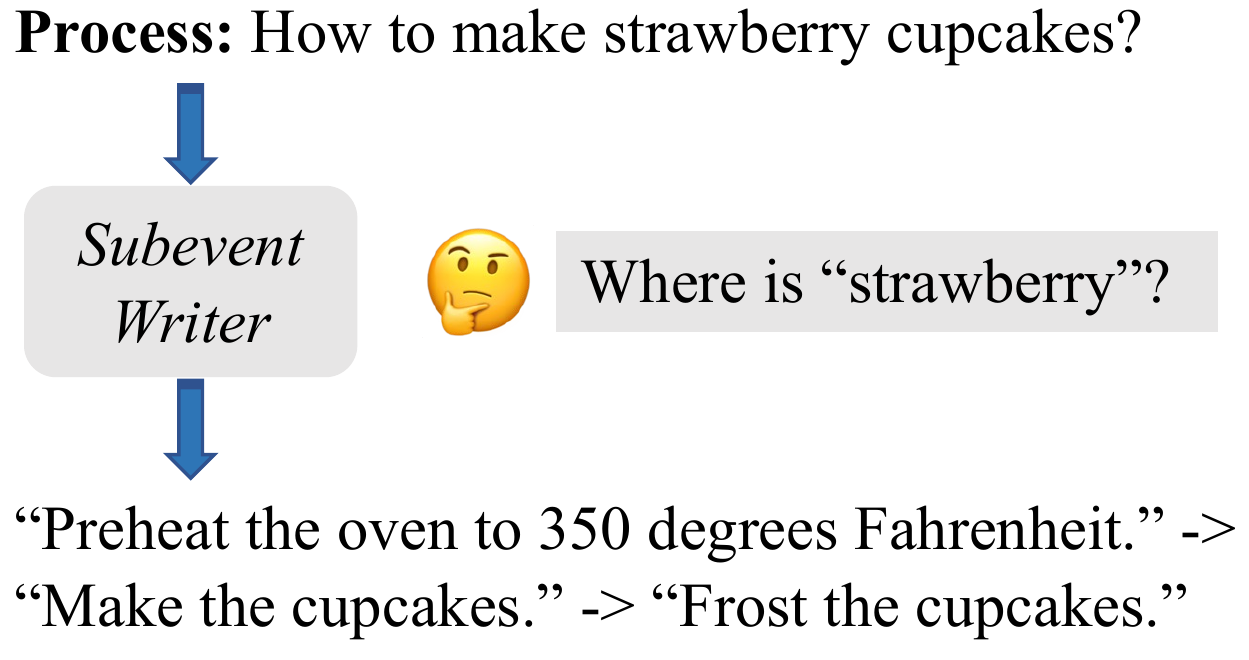}
    \caption{An error analysis. The \emph{SubeventWriter} ignores ``strawberry'' in the question and answers how to make cupcakes.}
    \label{error_analysis_figure}
\end{figure}

The main limitation is that our \emph{SubeventWriter} framework lacks knowledge while it needs to understand multiple entities in processes and how they interact. As shown in Figure~\ref{error_analysis_figure}, we ask the framework ``How to make strawberry cupcakes?'' However, \emph{SubeventWriter} ignores ``strawberry'' in the question, which shows \emph{SubeventWriter} does not have knowledge about general cupcakes and strawberry cupcakes. Thus, it cannot infer the way to make strawberry cupcakes from making cupcakes. Future work can investigate effective ways to integrate more knowledge and give models stronger reasoning ability. For example, \newcite{zhang2020analogous} utilized the hierarchical structure among events to conceptualize and instantiate similar processes.

We also test human performance to show the limitations of \emph{SubeventWriter} and the large room for improvements. Notice that a process usually owns multiple ground truth references, which are annotated by humans. For every process, we randomly select a sub-event sequence from ground truth as a human prediction. The randomly selected one will be excluded from references.

From the results in Table~\ref{human_performance_table}, we can observe that there is still a notable gap between machine performance and human performance. For example, the BLEU-2 of human performance is more than twice of \emph{SubeventWriter} (T5-3b) (28.33\% vs. 11.30\%).

\begin{table}[t]
    \small
	\centering
	\setlength{\tabcolsep}{4pt}
	\begin{tabular}{l|cccc}
	    \toprule
		\textbf{Models}&\textbf{B-1}&\textbf{B-2} &\textbf{R-L}&\textbf{BERT}\\
		\midrule
        \emph{SubeventWriter} (T5-3b) &34.75&11.30&24.17&65.67\\
        \midrule
        Human Performance &54.47&28.33&38.28&73.18\\
		\bottomrule
	\end{tabular}
	\caption{Human performance on the WikiHow dataset. We add \emph{SubeventWriter} (T5-3b), which achieves the best machine performance.}
	\label{human_performance_table}
\end{table}



\bibliographystyle{acl_natbib}

\newpage
\appendix

\section{Input and Output Examples}
\label{appendix_io_example}
We list examples of input and output in this section, 

\subsection{Examples of the Prompt Template}
\label{appendix_prompt_io_example}
In Table~\ref{prompt_example1_table} and Table~\ref{prompt_example2_table}, we give two examples to show how we train \emph{SubeventWriter} with the prompt template. We train \emph{SubeventWriter} to generate one sub-event each time in chronological order and append it back to the input. If all sub-events are generated, \emph{SubeventWriter} generates ``none''.

\begin{table}[t]
    \small
	\centering
	\begin{tabular}{l|p{0.8\columnwidth}}
	    \toprule
	    \multicolumn{2}{p{0.9\columnwidth}}{\textbf{Process:} cook eggs} \\
	    \multicolumn{2}{p{0.9\columnwidth}}{\textbf{Reference:} Place eggs in a pot of water. $\rightarrow$ Bring the water to a boil. $\rightarrow$ Turn off the heat and place the eggs in cold water.}\\
	    \midrule
		\multirow{2}{*}{\textbf{1}}&\textbf{Input:} How to cook eggs? Step 1: [M]\\
		&\textbf{Output:} Place eggs in a pot of water.\\
	    \midrule
		\multirow{2}{*}{\textbf{2}}&\textbf{Input:} How to cook eggs? Step 1: Place eggs in a pot of water. Step 2: [M]\\
		&\textbf{Output:} Bring the water to a boil.\\
		\midrule
		\multirow{2}{*}{\textbf{3}}&\textbf{Input:} How to cook eggs? Step 1: Place eggs in a pot of water. Step 2: Bring the water to a boil. Step 3: [M]\\
		&\textbf{Output:} Turn off the heat and place the eggs in cold water.\\
		\midrule
		\multirow{2}{*}{\textbf{4}}&\textbf{Input:} How to cook eggs? Step 1: Place eggs in a pot of water. Step 2: Bring the water to a boil. Step 3: Turn off the heat and place the eggs in cold water. Step 4: [M]\\
		&\textbf{Output:} none\\
		\bottomrule
	\end{tabular}
	\caption{An example of the process ``cook eggs'' with the prompt template. Each sub-event in the sequence is regarded as the output in turns. [M] is a masked token used in pre-train models, like <extra\_id\_0> of T5.}
	\label{prompt_example1_table}
\end{table}

\begin{table}[t]
    \small
	\centering
	\begin{tabular}{l|p{0.8\columnwidth}}
	    \toprule
	    \multicolumn{2}{p{0.9\columnwidth}}{\textbf{Process:} buy a house} \\
	    \multicolumn{2}{p{0.9\columnwidth}}{\textbf{Reference:} Getting your financials in order. $\rightarrow$ Shopping for a home. $\rightarrow$ Making an offer and finalizing the deal.}\\
	    \midrule
		\multirow{2}{*}{\textbf{1}}&\textbf{Input:} How to buy a house? Step 1: [M]\\
		&\textbf{Output:} Getting your financials in order.\\
	    \midrule
		\multirow{2}{*}{\textbf{2}}&\textbf{Input:} How to buy a house? Step 1: Getting your financials in order. Step 2: [M]\\
		&\textbf{Output:} Shopping for a home.\\
		\midrule
		\multirow{2}{*}{\textbf{3}}&\textbf{Input:} How to buy a house? Step 1: Getting your financials in order. Step 2: Shopping for a home. Step 3: [M]\\
		&\textbf{Output:} Making an offer and finalizing the deal.\\
		\midrule
		\multirow{2}{*}{\textbf{4}}&\textbf{Input:} How to buy a house? Step 1: Getting your financials in order. Step 2: Shopping for a home. Step 3: Making an offer and finalizing the deal. Step 4: [M]\\
		&\textbf{Output:} none\\
		\bottomrule
	\end{tabular}
	\caption{An example of the process “buy a house” with the prompt template. Each sub-event in the sequence is regarded as the output in turns. [M] is a masked token used in pre-train models, like <extra\_id\_0> of T5.}
	\label{prompt_example2_table}
\end{table}

\begin{table}[t]
    \small
	\centering
	\begin{tabular}{p{0.9\columnwidth}}
	    \toprule
	    \textbf{Process:} have a relaxing evening \\
	    \textbf{Reference:} Turn the lights down. $\rightarrow$ Put on some music or some relaxing nature sounds. $\rightarrow$ Make sure the temperature is comfortable. $\rightarrow$ Turn your phone off.\\
	    \midrule
		\textbf{Example:} How to have a relaxing evening? Step 1: Turn the lights down. Step 2: Put on some music or some relaxing nature sounds. Step 3: Make sure the temperature is comfortable. Step 4: Turn your phone off.\\
		\textbf{Label:} Positive\\
	    \midrule
		\textbf{Example:} How to have a relaxing evening? Step 1: Turn the lights down. Step 2: Put on some music or some relaxing nature sounds. \Blue{Step3: Turn the lights down.} Step 4: Make sure the temperature is comfortable. Step 6: Turn your phone off.\\
		\textbf{Label:} Negative (with a duplicate sub-event)\\
		\midrule
		\textbf{Example:} How to have a relaxing evening? Step 1: Turn the lights down. Step 2: Put on some music or some relaxing nature sounds. Step 3: Make sure the temperature is comfortable. \Blue{Step 4: Place eggs in a pot of water.} Step 5: Turn your phone off.\\
		\textbf{Label:} Negative (with an irrelevant sub-event)\\
		\bottomrule
	\end{tabular}
	\caption{Positive and negative examples of the process ``have a relaxing evening'' for the training stage of the coherence controller. We mark the sub-event used to build negative examples with blue color.}
	\label{training_coherence_example_table}
\end{table}

\subsection{Examples of Training the Coherence Controller}
\label{appendix_training_coherence_controller_example}
In Table~\ref{training_coherence_example_table}, we give positive and negative examples to show how we train the coherence controller. For negative examples, we provide examples for using a duplicate sub-event to corrupt the \textbf{\emph{local coherence}} and using an irrelevant sub-event to corrupt the \textbf{\emph{global coherence}}.

\subsection{Examples of the Inference Stage of the Coherence Controller}
\label{appendix_inference_coherence_controller_example}
In Table~\ref{inference_coherence_example_table}, we give two candidates for the third sub-event in the process ``make a felt hear card''. We also show the input to the coherence controller, which is concatenated from the process, sub-events generated in prior iterations, and current candidates. The coherence controller can assign coherent input higher scores and penalize incoherent input.

\begin{table}[H]
    \small
	\centering
	\begin{tabular}{l|p{0.8\columnwidth}}
	    \toprule
	    \multicolumn{2}{p{0.9\columnwidth}}{\textbf{Process:} make a felt heart card} \\
	    \multicolumn{2}{p{0.9\columnwidth}}{\textbf{Sub-events generated in prior iterations:} Cut out a heart shape. $\rightarrow$ Glue the heart to a card base.}\\
	    \midrule
		\multirow{3}{*}{\textbf{1}}&\textbf{Candidate:} Add a sentiment. \\
		&\textbf{Input:} How to make a felt heart card? Step 1: Cut out a heart shape. Step 2: Glue the heart to a card base. Step 3: Add a sentiment.\\
		&\textbf{Coherence Score:} 0.23 (low score)\\
	    \midrule
		\multirow{3}{*}{\textbf{1}}&\textbf{Candidate:} Add embellishments. \\
		&\textbf{Input:} How to make a felt heart card? Step 1: Cut out a heart shape. Step 2: Glue the heart to a card base. Step 3: Add embellishments.\\
		&\textbf{Coherence Score:} 0.82 (high score)\\
		\bottomrule
	\end{tabular}
	\caption{An example of the process ``make a felt heart card'' for the inference stage of the coherence controller. We compare two candidates for the third sub-events. We can see that the coherence controller can score the coherent candidate higher.}
	\label{inference_coherence_example_table}
\end{table}

\section{Implementation Details}
We conduct all experiments on 8 NVIDIA A100 GPUs.
\label{appendix_implementation}
\subsection{Coherence Controller}
The coherence controller is fine-tuned on BERT-base due to efficiency. We also tested three other variants of the Transformer~\cite{vaswani2017attention}: BERT-large, RoBERTa-large~\cite{liu2019roberta}, and RoBERTa-base. We fine-tune them with sub-event sequences from WikiHow to keep the domain consistent inside \emph{SubeventWriter}. Two negative examples are sampled using \textbf{\emph{duplicate sub-event}} and the same number using \textbf{\emph{irrelevant sub-event}} ($2N=4$ in total).

We build two testing sets with positive and negative samples of 1:1. Negative examples in the first testing set are examples with corrupted local coherence, while those in the second set are examples with corrupted global coherence. Accuracy is shown in Table~\ref{coherence_model_performance} on both testing sets. We also show accuracy on all testing data of both sets (``All''). We can observe that BERT-base already achieves satisfying accuracy (93.14\%). Using larger models does not improve too much and increases computation cost.
\begin{table}[t]
	\small
	\centering
	\begin{tabular}{l|ccc}
	    \toprule
		\textbf{Models}&\textbf{Local}&\textbf{Global} &\textbf{All}\\
		\midrule
		BERT-base & 95.76 & 90.52 & 93.14\\
		BERT-large & 95.57 & 91.62 & 93.59\\
		RoBERTa-base & 97.11 & 92.16 & 94.63 \\
		RoBERTa-large & 96.63 & 94.17 & 95.40 \\
		\bottomrule
	\end{tabular}
	\caption{Accuracy of coherence controllers. ``Local'' and ``Global'' refer to testing sets with corrupted local and global coherence, respectively. ``All'' contains all testing data of both sets.}
	\label{coherence_model_performance}
\end{table}

\subsection{Best Hyper-parameters}
\label{appendix_hyper}
We collect the best hyper-parameters of \emph{SubeventWriter} and All-at-once Seq2Seq in Table~\ref{hyper_parameter_table}, including learning rate, batch size, and the weight $\lambda$ of coherence scores (Eq.~\ref{sum_score_equation}).

\begin{table}[t]
	\small
	\centering
	\begin{tabular}{l|ccc|cc}
	    \toprule
		\multirow{2}{*}{\textbf{Models}}& \multicolumn{3}{c|}{\textbf{Ours}} & \multicolumn{2}{c}{\textbf{All-at-once}}\\ \cmidrule(lr){2-4} \cmidrule(lr){5-6}
		& LR & BS &\textbf{$\lambda$}& LR & BS \\
		\midrule
        BART-base & 5e-5 & 32 & 2 & 5e-5 & 32\\
        BART-large & 5e-5 & 32 & 1 & 1e-5 & 32\\
		\midrule
        T5-base & 5e-4 & 64 & 5 & 1e-3 & 32\\
        T5-large & 5e-5 & 32 & 0.5 & 5e-4 & 32\\
        T5-3b & 5e-5 & 64 & 0.5 & 1e-4 & 32\\
		\bottomrule
	\end{tabular}
	\caption{The best hyper-parameters for \emph{SubeventWriter} (``Ours'') and All-at-once Seq2Seq (``All-at-once''). ``LR'', ``BS'' and ``$\lambda$'' refer to learning rate, batch size and the weight $\lambda$, respectively.}
	\label{hyper_parameter_table}
\end{table}

\begin{table*}[t]
    \small
	\centering
	\begin{tabular}{l||cccc|cc}
	    \toprule
		\textbf{Models}&\textbf{B-1}&\textbf{B-2} &\textbf{R-L}&\textbf{BERT} & \textbf{$\Delta_{B\mhyphen1}$} & \textbf{$\Delta_{B\mhyphen2}$}\\
		\midrule
		Zero-shot Large LM (GPT-J) &14.02&0.42&16.53&45.33 & - & - \\
		Zero-shot Large LM (T5-11b) &20.31&0.97&14.13&54.72 & - & - \\
		\midrule
		Top-1 Similar Sequence (Glove) &16.74&1.08&11.85&57.55& - & -\\
		Top-1 Similar Sequence (\textsc{SBert}) &18.12&1.86&13.06&59.95& - & -\\
		\midrule
        All-at-once Seq2Seq (BART-base) &21.09&4.41&18.70&58.84& - & -\\
        All-at-once Seq2Seq (BART-large) &22.39&4.77&19.09&59.50& - & -\\
		\midrule
        All-at-once Seq2Seq (T5-base) &20.51&5.52&19.76&51.83& - & -\\
        All-at-once Seq2Seq (T5-large) &24.39&7.30&21.64&57.23& - & -\\
        All-at-once Seq2Seq (T5-3b) &28.22&8.60&22.98&62.08& - & -\\
        \midrule
        \emph{SubeventWriter} (BART-base) &29.70&8.26&21.33&60.26 & $\uparrow$ 8.61 & $\uparrow$ 3.85\\
        \emph{SubeventWriter} (BART-large) &31.57&9.44&22.17&61.95& $\uparrow$ 9.18 & $\uparrow$ \textbf{4.67}\\
        \midrule
        \emph{SubeventWriter} (T5-base) &32.09&9.31&22.51&61.99 & $\uparrow$ \textbf{11.58} & $\uparrow$ 3.79\\
        \emph{SubeventWriter} (T5-large) &33.97&10.65&23.33&64.40& $\uparrow$ 9.58 &
        $\uparrow$ 3.35\\
        \emph{SubeventWriter} (T5-3b) &\textbf{35.64}&\textbf{12.07}&\textbf{24.08}&\textbf{65.79}& $\uparrow$ 7.42 & $\uparrow$ 3.47\\
		\bottomrule
	\end{tabular}
	\caption{Performance of all frameworks on the validation data of the WikiHow dataset.}
	\label{wikihow_valid_table}
\end{table*}

\section{Results on WikiHow Validation Dataset}
\label{appendix_wikihow_valiation}
We collect the performance on the validation set of the WikiHow dataset in Table~\ref{wikihow_valid_table}. \emph{SubeventWriter} also works well on validation data.

\section{Main Evaluation and Analysis}
We provide complementary results of main evaluation and analysis as follows.
\subsection{Full Results of Ablation Study}
\label{appendix_ablation_study}
Here we present the ablation study results of \emph{SubeventWriter} based on all BART and T5 models in Table~\ref{BART_ablation_study_table} and Table~\ref{T5_ablation_study_table}, respectively.

\begin{table}[H]
    \small
	\centering
	\begin{tabular}{l|cccc}
	    \toprule
		\textbf{Models}&\textbf{B-1}&\textbf{B-2}&\textbf{R-L}&\textbf{BERT}\\
		\midrule
		Ours (BART-base) &\textbf{29.62}&\textbf{8.35}&\textbf{21.59}&\textbf{60.42}\\
		\midrule
        $\diamond$ w/o CoCo &25.82&7.18&21.24&56.60\\
		$\diamond$ w/o CoCo \& ITER. &21.01&4.52 &18.83&58.79\\
		\midrule
		\midrule
		Ours (BART-large) &\textbf{31.31}&\textbf{9.41}&\textbf{22.52}&\textbf{61.83} \\
		\midrule
        $\diamond$ w/o CoCo &26.78&7.79&22.04&59.53\\
		$\diamond$ w/o CoCo \& ITER. &21.84&4.73 &18.94&59.45\\
		\bottomrule
	\end{tabular}
	\caption{Ablation study results on BART-base and BART-large.}
	\label{BART_ablation_study_table}
\end{table}

\subsection{Full Results of Few-shot Learning}
\label{appendix_few_shot}
We offer full results of few-shot learning on the testing set of WikiHow dataset in Table~\ref{allatonce_few_shot_table} for All-at-once Seq2Seq and Table~\ref{subeventwriter_few_shot_table} for \emph{SubeventWriter}.

\begin{table}[H]
    \small
	\centering
	\begin{tabular}{l|cccc}
	    \toprule
		\textbf{Models}&\textbf{B-1}&\textbf{B-2}&\textbf{R-L}&\textbf{BERT}\\
		\midrule
		Ours (T5-base) &\textbf{30.74}&\textbf{8.89}&\textbf{22.44}&\textbf{61.81}\\
		\midrule
        $\diamond$ w/o CoCo &28.56&8.27&22.01&58.89\\
		$\diamond$ w/o CoCo \& ITER. &20.33&5.63& 20.22&52.15\\
		\midrule
		\midrule
		Ours (T5-large) &\textbf{33.01}&\textbf{10.39}&\textbf{23.07}&\textbf{64.19}\\
		\midrule
        $\diamond$ w/o CoCo &30.41&9.14&22.75&62.19\\
		$\diamond$ w/o CoCo \& ITER. &24.27&7.11&21.76&57.58\\
		\midrule
		\midrule
		Ours (T5-3b) &\textbf{34.75}&\textbf{11.30}&\textbf{24.17}&\textbf{65.67}\\
		\midrule
        $\diamond$ w/o CoCo &33.63&10.90&24.15&65.57\\
		$\diamond$ w/o CoCo \& ITER. &27.99&8.72&23.36&62.03\\
		\bottomrule 
	\end{tabular}
	\caption{Ablation study results on T5-base, T5-large and T5-3b.}
	\label{T5_ablation_study_table}
\end{table}

\begin{table}[H]
    \small
	\centering
	\begin{tabular}{p{1.5cm}|l|cccc}
	    \toprule
		\textbf{All-at-once} & \textbf{\textsc{\#Shot}}&\textbf{B-1}&\textbf{B-2}&\textbf{R-L}&\textbf{BERT}\\
		\midrule
		\multirow{4}{*}{T5-base} & n = 5K &16.37&3.22&16.78&47.65\\
        &n = 10K &16.60&3.71&17.44&48.50\\
		&n = 20K &18.43&4.31&18.30&49.90\\
	    &n = 30K &18.65&4.72&18.54&50.22\\
	    \midrule
		\multirow{4}{*}{T5-large} & n = 5K &20.15&4.62&19.07&54.27\\
        &n = 10K &20.59&5.09&19.69&55.21\\
		&n = 20K &22.35&5.86&20.45&56.93\\
	    &n = 30K &23.00&6.49&21.06&56.52\\
		\bottomrule 
	\end{tabular}
	\caption{Few-shot learning results of All-at-once Seq2Seq based on T5-base and T5-large on the testing set.}
	\label{allatonce_few_shot_table}
\end{table}

\subsection{Full Results of T5-small}
\label{appendix_t5_small}
We show the full results of \emph{SubeventWriter} (T5-small) and All-at-once Seq2Seq (T5-small) in Table~\ref{t5_small_result_table}. The performance drops compared to \emph{SubeventWriter} (T5-base) and All-at-once Seq2Seq (T5-base), respectively, as shown in Table~\ref{t5_small_change_table}.
Notice that we test the \emph{SubeventWriter} without the coherence controller. Thus, to calculate performance changes, please refer to ``$\diamond$ w/o CoCo'' of \emph{SubeventWriter} (T5-base) in the ablation study (Table~\ref{BART_ablation_study_table} and Table~\ref{T5_ablation_study_table}).

\begin{table}[H]
    \small
	\centering
	\begin{tabular}{p{1.5cm}|l|cccc}
	    \toprule
		\textbf{Ours} & \textbf{\textsc{\#Shot}}&\textbf{B-1}&\textbf{B-2}&\textbf{R-L}&\textbf{BERT}\\
		\midrule
		\multirow{4}{*}{T5-base} & n = 5K &23.68&5.30&19.71&54.77\\
        &n = 10K &24.00&5.43&19.77&55.29\\
		&n = 20K &25.47&6.02&20.43&56.45\\
	    &n = 30K &25.49&6.17&20.57&56.98\\
	    \midrule
		\multirow{4}{*}{T5-large} & n = 5K &29.07&7.41&21.48&62.24\\
        &n = 10K &29.14&7.60&21.99&62.43\\
		&n = 20K &30.23&8.73&22.28&63.42\\
	    &n = 30K &30.67&8.85&22.73&63.42\\
		\bottomrule 
	\end{tabular}
	\caption{Few-shot learning results of \emph{SubeventWriter} based on T5-base and T5-large on the testing set.}
	\label{subeventwriter_few_shot_table}
\end{table}

\begin{table}[H]
    \small
	\centering
	\begin{tabular}{l|l|cccc}
	    \toprule
		\textbf{Split}&\textbf{Models}&\textbf{B-1}&\textbf{B-2}&\textbf{R-L}&\textbf{BERT}\\
		\midrule
		\multirow{2}{*}{valid} &All-at-once &17.80&4.28&18.27&47.96\\
        &\emph{SubeventWriter} &\textbf{22.98}&\textbf{5.47}&\textbf{19.58}&\textbf{51.38}\\
		\midrule
		\multirow{2}{*}{test} &All-at-once &17.18&4.25&17.90&48.26\\
        &\emph{SubeventWriter} &\textbf{23.53}&\textbf{5.67}&\textbf{19.69}&\textbf{51.26}\\
		\bottomrule
	\end{tabular}
	\caption{Performance of using T5-small on validation and testing sets. ``valid'' and ``test'' are shortened forms of validation and testing.}
	\label{t5_small_result_table}
\end{table}

\begin{table}[H]
    \small
	\centering
	\begin{tabular}{l|l|cccc}
	    \toprule
		\textbf{Split}&\textbf{Models}&\textbf{$\delta_{B\mhyphen1}$}&\textbf{$\delta_{B\mhyphen2}$}&\textbf{$\delta_{R\mhyphen L}$}&\textbf{$\delta_{BERT}$}\\
		\midrule
		\multirow{2}{*}{valid} &All-at-once & 2.71 & 1.24 & 1.49 & 3.87\\
        &\emph{SubeventWriter} & 6.76 & 3.12 & 2.64 & 7.48\\
		\midrule
		\multirow{2}{*}{test} &All-at-once & 3.15 & 1.38 & 2.32 & 3.89\\
        &\emph{SubeventWriter} & 5.03 & 2.60 & 2.32 & 7.63\\
		\bottomrule
	\end{tabular}
	\caption{Performance drops when we replace T5-base with T5-small.}
	\label{t5_small_change_table}
\end{table}

\end{document}